\documentclass[10pt,twocolumn,letterpaper]{article}

\usepackage{cvpr}              

\usepackage{soul}
\usepackage{multirow}

\newcommand{\Major}[1]{{}}
\newcommand{\KAnote}[1]{{\color{teal}{\bf KA: }#1}} 
\newcommand{\KA}[1]{{\color{black}#1}} 
 \newcommand{\KGnote}[1]{{\color{blue}{\bf KG: }#1}}
\newcommand{\KG}[1]{{\color{magenta}#1}} 
\newcommand{\JW}[1]{{\color{orange}#1}} 
\newcommand{\JWnote}[1]{{\color{ForestGreen}{\bf JW: }#1}}

\newcommand{\KAnew}[1]{{\color{brown}#1}} 
\renewcommand{\JW}{\textcolor{black}}
\renewcommand{\KG}{\textcolor{black}}
\renewcommand{\KAnew}{\textcolor{black}}

\newcommand{\JWnew}[1]{{\color{red}#1}}
\renewcommand{\JWnew}{\textcolor{black}}

\renewcommand{\JWnote}[1]{{}}
\renewcommand{\KAnote}[1]{{}}

\newcommand{\JWCam}[1]{{\color{black}#1}}

\usepackage{amsmath}
\DeclareMathOperator*{\argmax}{arg\,max}

\usepackage{array}
\newcommand{\PreserveBackslash}[1]{\let\temp=\\#1\let\\=\temp}
\newcolumntype{C}[1]{>{\PreserveBackslash\centering}p{#1}}
\newcolumntype{L}[1]{>{\PreserveBackslash\raggedright}p{#1}}

\newcommand{\modelname}{Stitch-a-\KAnew{Demo}}

\usepackage{tcolorbox}
\tcbuselibrary{breakable} 

\usepackage[symbol*]{footmisc}

\definecolor{cvprblue}{rgb}{0.21,0.49,0.74}
\usepackage[pagebackref,breaklinks,colorlinks,allcolors=cvprblue]{hyperref}

\def\paperID{XXXX} 
\def\confName{CVPR}
\def\confYear{2026}

\title{\modelname: \KG{Creating} Video Demonstrations from Multistep Descriptions}

\author{Chi Hsuan Wu$^{*}$, Kumar Ashutosh$^{*}$, Kristen Grauman\\
UT Austin\\
}

\begin{document}

\twocolumn[{%
\renewcommand\twocolumn[1][]{#1}%
\maketitle

\begin{center}
\vspace*{-0.25in}
  \centering
    \includegraphics[width=\linewidth]{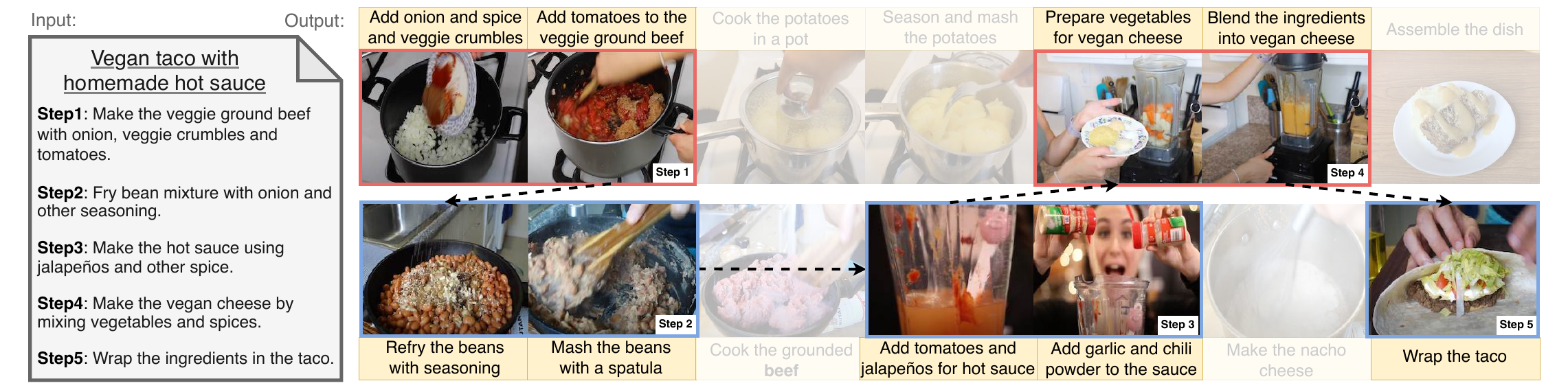} 
    \vspace{-0.2in}
    \captionof{figure}{\textbf{Video demonstration from multistep descriptions}. Given multistep descriptions (left) 
    aiming to achieve a procedural task, \eg making \emph{vegan taco}, our method obtains clips from thousands of 
    instructional videos to visually demonstrate the \KAnew{procedure} (right). The goal is for every clip to correctly describe a step, while maintaining visual consistency. 
    Our proposed method goes beyond current retrieval and generation methods, 
    which fail to faithfully ground multistep procedures in coherent visual sequences.
    }
  \label{fig:teaser-fig}
\end{center}%
}]

\begin{abstract}

\vspace{-0.4cm}

When obtaining visual illustrations 
from text descriptions, 
today's methods take a description with a single text context---a caption, or an action description---and  retrieve or generate the matching visual context. However, prior work does not 
permit visual 
illustration of
\emph{multistep} descriptions, \eg a cooking recipe \JW{or a gardening instruction manual}
, and  
simply handling each step description in isolation would result in an incoherent demonstration.  
We propose \modelname, a novel retrieval-based method to assemble a 
video demonstration from a multistep description. The resulting video contains clips, possibly from different sources, that accurately reflect all the step descriptions, while being visually coherent.
We formulate a
training pipeline that creates large-scale weakly supervised data containing diverse 
\KAnew{procedures} 
and 
injects hard negatives that promote both correctness and coherence.
Validated on in-the-wild instructional videos, \modelname~achieves
state-of-the-art performance, with 
gains up to 29\% as well as dramatic wins
in a human preference study. 

\footnotetext[1]{Equal contribution.} \footnotetext{Project page: \href{https://vision.cs.utexas.edu/projects/stitch-a-demo/}{https://vision.cs.utexas.edu/projects/stitch-a-demo/}}

\end{abstract}


\vspace*{-0.05in}
\section{Introduction}
\label{sec:intro}

Instructional or ``how-to'' videos are commonly used to learn new skills---such as cooking, \KAnew{gardening}, repairing bikes, or yoga. These videos contain an explanation of the task, often in the form of multiple procedural \emph{steps}, along with a rich visual demonstration. 
Video demonstrations have shown to be a great learning aid \cite{surgenor2017impact}, significantly augmenting written step descriptions for human learners.  \KG{Meanwhile, in robot learning, training with ``passive" video of human skill executions is increasingly attractive for representation learning and efficient imitation~\cite{SeeDo, human_robot}.}



Despite the scale of instructional videos on the internet, they still only represent a sliver of all possible demonstration sequences, given the combinatorics of how different steps can potentially be combined. 
Any given video assumes a fixed sequence of steps, which might differ from the step description that a person wants to visualize, whether from a recipe book\KAnew{, an instructional manual,} or their own imagination. 
For example, 
consider the step sequence
to prepare a \emph{Vegan Mexican Taco} by (a) making vegan ground beef, (b) mixing onions and seasoning, (c) making sauce, (d) making vegan cheese, and (e) wrapping it up.  
What if none of the videos on making a \emph{Mexican Taco} covers these exact steps, in order? Some recipe might add mashed potatoes (Fig.~\ref{fig:teaser-fig}, top row) or make nacho cheese (Fig.~\ref{fig:teaser-fig}, bottom row). 
The problem is that no one video demonstration may show
the
\emph{exact} steps of interest. 
The task to provide
a faithful video demonstration for a given sequence of textual step descriptions is technically challenging.  Not only is the space of possible \JW{procedures} very large, but also the wide range of expertise, availability of the tools and objects, and the multistep  dependencies of procedural actions all add to the challenge. 
On the one hand, prior text-to-video methods~\cite{covr,detours,internvideo,hiervl,egovlp,mil-nce,videoclip,hero,internvideo2} 
can retrieve clips
that achieve good instantaneous video-text alignment capturing semantic similarity, but they stop short of retrieving consistent multistep depictions.
On the other hand, video or image generation~\cite{emu-video,genhowto,showhowto,GenIllustrated} 
shines for creating imaginative outputs beyond the boundaries of any given dataset, but suffers from 
high computational requirements and hallucinations that  reduce realism.




We propose \modelname, a novel method to obtain video demonstrations from multistep descriptions. We \emph{retrieve} clips from multiple videos that best satisfy the input step description 
while ensuring 
\KG{temporal}, visual, and environment\KAnew{al} consistency. 
Our approach employs a novel \emph{\KAnew{procedure} evaluator} network, together with well-designed positives and negatives for training that represent correctness and visual consistency constraints. The training method leverages the prior from strong instantaneous video-text signals, and combines them with our designed constraints, to obtain correct video demonstrations from multistep descriptions. Moreover, we propose an adaptive search space reduction for real applications where the search space is large.

We consider three diverse procedural domains.  We focus on cooking, \KG{which not only represents the single most popular procedural activity in online how-to's today,} but also is 
particularly compelling due to its diversity of tools, ingredients, and physical techniques.
\KG{Indeed, many influential large datasets and models center around cooking~\cite{howto100m,epic-kitchens-100,epic-fields,epic-sounds,htstep-neurips2023,detours,stepdiff,gepsan,videoosc,recipe2video}.
To underscore the generality of our approach, } we 
further translate the same model to other instructional domains---woodworking and gardening---that also exhibit 
multiple valid materials and techniques for completing a task.
We introduce a large-scale weakly-supervised \modelname~training dataset and a manually curated testing dataset.
Rigorous quantitative experiments with in-the-wild videos and a human preference study show the effectiveness of our \modelname~over state-of-the-art visual demonstration generation and retrieval methods~\cite{detours,covr,showhowto,recipe2video}.




\section{Related Work}
\label{sec:related}

\textbf{Video and language representation learning.} Videos are often paired with text captions or narrations that can help learn strong associations between text and video~\cite{howto100m,egovlp,hiervl,egovlpv2,videoclip}. These video representations can then be used for a variety of downstream tasks, including captioning \cite{captioning-1,captioning-2,captioning-3,captioning-4,captioning-5}, text-to-video and video-to-text retrieval \cite{epic-kitchens-100,howto100m,videoclip,mil-nce,vid-retrieval-1,vid-retrieval-2,vid-retrieval-3,univl}, and action recognition \cite{omnivore,mvitv2,uniformer,memvit,slowfast}. Most of these tasks consider clips that are typically a few seconds long. Other tasks requiring longer video understanding are action anticipation~\cite{avt,hiervl,ego4d} and procedure planning \cite{chang2020procedure,procedure2,procedure3,p3iv,pdpp}.
However, the query or description in these tasks is a single sentence. 
Unlike prior work, we learn to \JWCam{provide a} 
video 
\JWCam{given} multistep descriptions, effectively learning associations over long videos with procedural steps.


\textbf{Learning from instructional videos.} Beyond their use in video representation learning, instructional videos are useful for step understanding~\cite{coin,crosstask,video-distant,task_graph}, procedural planning~\cite{procedure3,procedure2,chang2020procedure,p3iv,pdpp}, 
and step grounding~\cite{tan,vina,drop-dtw}. Due to the detailed step descriptions \KG{accompanying visual demos} in instructional videos, 
these tasks enable a deep understanding of standard procedures.
Recent methods \cite{video-mined,paprika,detours} go beyond a single video demonstration and attempt to reason across multiple demonstrations for task graph learning~\cite{unitygraph,paprika,egoexo4d,video-distant,graph2vid,task-graph-objective,task_graph}. 
Despite learning from multiple videos, 
during inference those methods still 
handle one video at a time, \eg to perform step 
forecasting. In contrast, we learn to
stitch clips from multiple videos into a cohesive and correct visual demonstration,
unlocking a deeper understanding of instructional videos.


\textbf{Video or image from text descriptions.} Obtaining a video or image from text has been studied mostly from two approaches---generation and retrieval. Media generation \cite{emu-video,instructpix2pix,hu2024instruct} is used to create \emph{any} image or video 
from text descriptions---even unrealistic ones. Controlled media generation \cite{instructpix2pix,lin2024text,tu2024motioneditor} is also used to edit images and videos to incorporate the desired change. 
\JW{With current generative model capabilities, video generation remains limited to short clips~\cite{moviegen, sora, cogvideox}. For instructional content (typically 5-30 minutes), prior works therefore generate a single illustrative image per step~\cite{GenIllustrated, genhowto, showhowto, lego}, showing limited action information. Media generation is also prone to hallucinations, often producing unrealistic step illustrations unlike retrieval.}
We show that human judges prefer our method compared to the state-of-the-art ShowHowTo for image generation~\cite{showhowto}. 




\begin{figure*}[t] 
    \centering
    \includegraphics[width=\textwidth]{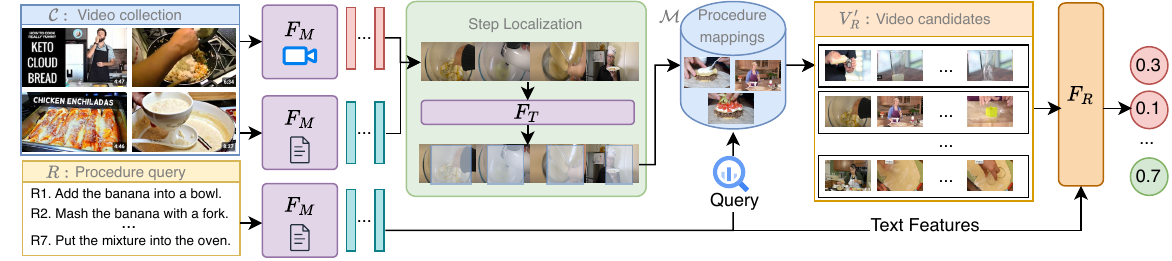} 
    \vspace{-0.5cm}
    \caption{\textbf{Method overview.} The videos and the step descriptions in $\mathcal{C}$ are used to create a \JW{procedure} mapping $\mathcal{M}$, using step localization~$F_T$. The \JW{procedure} query $R$ and $\mathcal{M}$ give video candidates \KAnew{$V'_R$}. The \emph{\JW{procedure} evaluator} $F_R$ outputs the likelihood of each candidate.}
    \label{fig:overview}
    \vspace{-0.3cm}
\end{figure*}

Video retrieval is the preferred approach when the right answer is known to exist in the candidate set. For instructional videos, retrieval has been used extensively in prior work~\cite{epic-kitchens-100,howto100m,videoclip,mil-nce,vid-retrieval-1,vid-retrieval-2,vid-retrieval-3,univl, Hirest}. Beyond the standard text-to-video retrieval setting, CoVR~\cite{covr} 
retrieves a video demonstration based on a reference video/image and a modification text. 
\KAnew{Limited prior work explores video
retrieval to illustrate cooking tasks~\cite{detours} or recipes~\cite{recipe2video}}
(and inversely inferring a recipe from a photo~\cite{romero}).  
VidDetours~\cite{detours} 
identifies a detour between two cooking videos 
using a user's language query, e.g., ``how do I make this without a blender?", a problem that is interesting but distinct from illustrating a sequence of step descriptions. Recipe2Video~\cite{recipe2video} 
creates a slideshow of each step and its image/video/audio demonstration, but is limited to retrieving clips based on rigid and inflexible metrics
which impedes correctness, coherence, and object state consistency, as we show in results.

\KAnew{None
of the existing retrieval methods is capable of retrieving visually and logically coherent video demonstrations from sequential step descriptions, as we tackle in this work.
Furthermore, unlike~\cite{recipe2video,detours,romero}, we go beyond cooking to demonstrate our method on domains like gardening and woodworking.
}



\vspace*{-0.05in}
\section{Method}

We first formally introduce the task in Sec.~\ref{subsec:task}, followed by the model design in Sec.~\ref{sec:model}, the dataset construction idea in Sec.~\ref{sec:dataset}, and finally the implementation details in Sec.~\ref{sec:implementation}.


\label{sec:formatting}


\subsection{Task formulation}
\label{subsec:task}

Given a multistep description for an instructional task, a.k.a.~a \KAnew{procedure or a recipe}, $R = (r_1, r_2, ..., r_n)$, where $r_i$ is a natural language step description (Fig. \ref{fig:overview} bottom left), and a collection of videos $\mathcal{C}~=~\{V^{(1)}, V^{(2)}, ..., V^{(N)}\}$ (Fig.~\ref{fig:overview} top left), we want to learn a function $\mathcal{F}$ that finds a video demonstration visually depicting the \KAnew{procedure} $R$. The output video $V_R~=~(v_1, v_2, ..., v_n)$, is a 
sequence of video clips, where $v_i$ is a segment from any video $V^{(j)}$, 
from time instances $t_1$ to $t_2$, \ie $v_i = V^{(j)}[t_1:t_2]$ and $V^{(j)} \in \mathcal{C}$, $t_1<t_2$. Overall, $\mathcal{F}(R, \mathcal{C}) = V_R$. \KAnew{To recall, videos in $\mathcal{C}$ have diverse human-object interactions~\cite{action-scene-graph}, object state changes~\cite{vidosc}, and step dependencies~\cite{task_graph}, making the task of learning $\mathcal{F}$ both interesting and challenging.}


\Major{\KGnote{this needs more rigor and more clear definition.} \KAnote{is this more than the notation change above?}}

Owing to the scale and the diversity of instructional videos on the internet~\cite{howto100m,youcook2,crosstask}, we create $V_R$ from multiple video demonstrations in $\mathcal{C}$ such that all the \JW{procedural} steps are correctly shown, with maximum visual consistency. 
Compared to applying image generation to illustrate a step~\cite{showhowto,GenIllustrated,genhowto,lego}, the proposed design has multiple advantages: it yields complete \emph{video} demonstrations, known to be more useful for human learning~\cite{surgenor2017impact}; it accounts for visual dependencies between the illustrated steps; and it is less prone to hallucinations and unrealistic outputs. 



Clips in $V_R$ 
need not originate from the same video: $v_i, v_j \in V_R, v_i \in V^{(x)}, v_j \in V^{(y)} \nRightarrow x=y$. We allow clips from different videos in the collection $\mathcal{C}$ to create $V_R$, since a single video demonstration may not be sufficient to represent an arbitrary \KAnew{procedure} $R$ (see Fig.~\ref{fig:teaser-fig}). 
Furthermore, the optimal $V_R$ should stitch together video clips that are not only \emph{correct} (demonstrate the target steps) but also \emph{coherent} (mutually consistent in terms of visual continuity and logic).  Our model accounts for both, as we detail next.

\subsection{\modelname~model design }
\label{sec:model}

Next we discuss the model design to learn $\mathcal{F}$. The high-level idea is to first temporally localize step descriptions in all videos to form clip and description tuples, \ie $(v, r)$, followed by creating a candidate set for training. We design a \emph{\KAnew{procedure} evaluator} module that determines the likelihood that a sequence of $(v_i, r_i)$, $i=1,\dots,n$, is a valid \KAnew{procedure}. This model is trained to obtain $V_R$ from $\{\mathcal{C}, R\}$. 
Fig.~\ref{fig:overview} shows the overview, and each part is described next.

\textbf{Encoding videos and \KAnew{procedure} text.} We use a multi-modal encoder $F_M$ (\eg~CLIP, InternVideo2 \cite{internvideo,internvideo2,clip}) to represent video and text. 
We obtain the \KAnew{procedure} text feature $\mathbf{r}_i = F_M(r_i)$ for a step $r_i$, and a video
\KA{clip feature $\mathbf{v}_i = F_M(v_i)$, where $v_i$ is a video segment for a \KAnew{procedure} step. \KAnew{We extract $1$ feature from $8$ frames per second. The video encodings are averaged over the duration of the video clip to obtain a step video clip feature, consistent with prior work~\cite{video-mined,video-distant,detours}.}
We do not train $F_M$.
}


\begin{figure*}[t] 
    \centering
    \includegraphics[width=\textwidth]{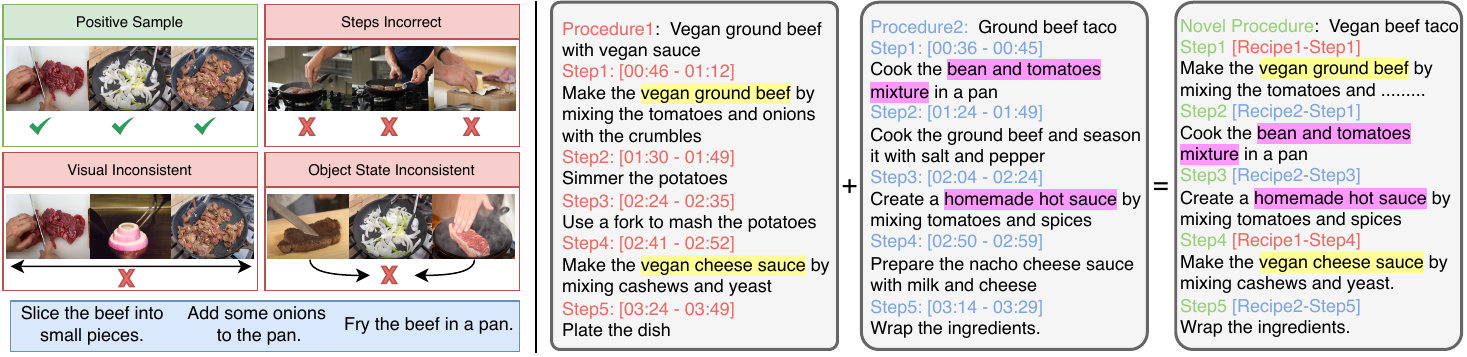} 
    \vspace{-0.5cm}
    \caption{\textbf{Examples of hard negatives and \JW{procedure} combination.} We design negative samples that violate step correctness, visual continuity, and object state continuity for contrastive learning (left). We show an example of combining step descriptions from $n$ (here $n=2$) video demonstrations into a novel \JW{procedure}, using an LLM~\cite{llama-adapter} (right).  The novel \JW{procedure} mixes steps from both descriptions. 
    }
    \label{fig:dataset_and_negatives}
    \vspace{-0.2cm}
\end{figure*}

\textbf{Localizing a clip in a video.} Retrieving clips for $V_R$ requires finding video clips associated with a \KAnew{procedural} step $r$.
We use a temporal localization function $F_T$ (\eg\cite{drop-dtw,vina,graph2vid,SVPTR}) 
to find the start and  end time of a step in a video. Specifically, \JW{let $R^{(j)}=(r'_1,r'_2,...r'_3)$ describe the procedure of $V^{(j)}$ in text. $[t_1, t_2] = F_T(V^{(j)}, r')$ which we use to obtain the clip $v' = V^{(j)}[t_1:t_2]$. This process yields a pool of procedure steps and clips $\mathcal{P}=\{(r',v')~|~r'\in R^{(j)}, v' \in V^{(j)}, V^{(j)} \in \mathcal{C}, r'\sim v'\}$. From $\mathcal{P}$ and the query $R$, we construct a procedure mapping $ \mathcal{M} = \{(r, v')~|~r \in R, (r',v')\in \mathcal{P}, r \sim r'\}$.
We use \KG{DropDTW~\cite{drop-dtw} as the} pre-trained $F_T$, and keep it frozen; any improvement there will only 
improve our model's performance.} 

\textbf{\emph{\KAnew{Procedure} evaluator} module.} \Major{\KGnote{consider if this is best name.}} The above modules help us obtain a map of \KAnew{a procedure's} steps and corresponding clips in the video collection $\mathcal{C}$. 
Next, we propose a \emph{\KAnew{procedure} evaluator} $F_R$ that finds the probability of the correctness of a candidate $V_R$, given the \JW{procedure} steps. Specifically, the \KAnew{procedure} correctness is given by 
\begin{displaymath}
    F_R\left((v'_1, v'_2, ..., v'_n)~\mid~v'_i \sim r_i, (r_i, v'_i) \in \mathcal{M}\right) \in [0, 1],
\end{displaymath}
where $v \sim r$ denotes the clip $v$ shows the step $r$ and $(v'_1, v'_2, ..., v'_n)$ is a full ``stitched" candidate sequence for $V_R$. $F_R$ has a transformer encoder that takes as input the concatenated features $F_M(r_i)$ and $F_M(v_i)$---one per \KAnew{procedural} step. We take the output of the transformer encoder corresponding to the CLS token, followed by a linear layer to output a probability, $F_R()$.

A key novelty of our method is how we train the \KAnew{procedure} evaluator. Instead of imposing heuristics to determine which are good stitched videos, our insight is to automatically generate a large-scale weakly supervised dataset with hard negatives that encourage correctness and coherence.

\textbf{Sampling negatives.}  
Building on ideas in contrastive learning~\cite{mil-nce,clip,videoclip}, 
we create a hard negative set by 
modifying correct video demonstrations in a targeted way. Each modification violates a constraint for correctness or visual coherence.  We describe each constraint first, followed by the modification we perform to obtain negative samples.

\begin{itemize}
    \item \textbf{Step and goal correctness}: All clips in $V_R$ must accurately represent the corresponding step description in $R$ and also contribute meaningfully to the goal of $R$ (and $V_R$). That is, $ v_i~\sim~r_i~\forall~i$. For example, if $r_i$ is \emph{``add salt to the chicken broth''}, 
    then $v_i$ should demonstrate sprinkling salt into the soup, but not performing other actions on the broth or adding salt to something else, \eg rice.

        \emph{Violating step and goal correctness}: Given a video demonstration $V_R$, we create a negative sample $V'_R$ by replacing any clip $v \in V_R, v \sim r$ with randomly selected $v'$ such that $v' \nsim r$. Furthermore, to make the negatives harder, we ensure the source video of $v'$ contributes to $V_R$, \ie $v' \in V^{(j)} \implies \exists v'' \in V^{(j)}, v'' \in V_R$. These constraints lead to a strong negative from the same video sources, violating the step correctness. See Fig.~\ref{fig:dataset_and_negatives} (left).

    \item \textbf{Visual continuity}: 
    We aim to 
    minimize the number of distinct video sources when constructing $V_R$. A new video must be selected only if the previously selected video is insufficient for a step $r_i$. Specifically, if $v_i~\in~V^{(j)}, v_i~\sim~r_i$, and there are two candidates for the next \JW{procedure} step, \ie $v \sim r_{i+1}, v' \sim r_{i+1}$, but $v \in V^{(j)}, v' \notin V^{(j)}$, then $v$ should be chosen as the next demonstration.  Note that many candidate subsequences will \emph{not} originate from the same video; hence we do have positives that cross video boundaries.
    

        \emph{Violating visual continuity}: If a video demonstration $V_R$ contains three consecutive steps from the same source video, we form a negative by replacing the middle clip with a similar demonstration from a different source video, while still ensuring step correctness. Specifically, if $v_i \in V_R, v_i \sim r_i, \text{~and~} v_i \in V^{(j)}~\forall i \in \{k, k+1, k+2\}$, we replace $v_{k+1}$ with $v' \notin V^{(j)}, v' \sim r_{k+1}$. See Fig.~\ref{fig:dataset_and_negatives}.

    
    \item \textbf{Object state continuity}: An object must not be in a state that has undergone an irreversible transformation at a previous time. For example, there should not be a step with a \emph{whole onion} after a step showing \emph{onions} being chopped, \KG{or a step showing \emph{unsanded wood} after a step showing the wood being sanded down.}

        \JWnew{\emph{Violating object state continuity}: We construct hard negatives in this category by 
        changing $v_x, v_y \in V_R$ to $v,v'\in V^{(j)}$, such that $v$ occurs before $v'$ in $V^{(j)}$, and $v \sim r_x, v'\sim r_y$, but $x>y$.} That is, the clips in the negative sample do not follow the temporal order in the original source video $V^{(j)}$. \KA{Even though some steps are interchangeable in \JW{procedures}, enforcing this constraint is helpful given the perceptual damage in ``undoing" a permanent transformation\KAnew{;} see Sec.~\ref{sec:results} and Supp.~\KAnew{where we experimentally validate the usefulness of violating object state continuity for training negatives.}}
        \Major{\KGnote{is this too blunt, meaning that we are trying to disallow any reversals, and yet not all of them will reflect visible object states; can we address how this won't cause harm or be too restrictive?  some steps are interchangeable.} \KAnote{Yes, but more often that not, it is not interchangable. Also helps improve the performance: ablation Tab~\ref{tab:ablations} (bottom) 4th row vs last.}}

\end{itemize}

In summary, we consider various realistic constraints when combining demonstrations from multiple \JW{procedures}. Our contrastive setup trains with these hard negatives. We show the effectiveness of each of these constraints in the results \KG{and discuss robustness to label noise below}.

\textbf{Training objective and inference.} A correct \KAnew{procedure} demonstration has the ground truth $F_R$ value of $1$; $0$ otherwise. During training, we use the standard Binary Cross Entropy (BCE) training objective. The choice is consistent with the output of the \JW{procedure} evaluator $F_R$ and the ground truth binary labels.

During inference, we have a set of video candidates $V'_R = (v'_1, v'_2, ..., v'_n)$ and the chosen video $V_R$ is 
\begin{displaymath}
    \argmax_{(v'_1, v'_2, ..., v'_n)} F_R(V'_R~|~v'_i \sim r_i).
\end{displaymath}
That is, the candidate with the highest probability from the \KAnew{procedure} evaluator.

\textbf{Adaptive search space reduction.} Finally, we discuss adaptive search space reduction for practical implementation. 
For a short clip or naive full-video retrieval, the candidate set scales as $O(N)$ for $N$ videos.
On the other hand, if video clips are allowed to be from distinct videos in an $M$-step \KAnew{procedure} with an average of $K$ clips in a video, the candidate set scales as $O((KN)^M)$.
Thus, we propose an adaptive search space reduction technique that can capture the ground truth in a much smaller candidate set, as follows.

We first construct a collection of sets $\mathcal{S}$ where an element is defined as 
\begin{displaymath}
    S_i := \{(x, v)~|~x\in \mathbb{Z}, v \in V^{(i)}, v \sim r_x \}.
\end{displaymath}
That is, each set contains \KAnew{procedural} step indices that have a matching video clip in the $i^{th}$ video. For example, if $V^{(1)}$ has clips $v_1$ and $v_2$ that match with steps $r_1$ and $r_2$ from the query $R$, then $S_1~=~\{(1, v_1), (2, v_2)\}$. \JWnew{This problem is the same as a set cover problem \cite{set-covering}. The task is to find a subset of $\mathcal{S}$, \ie a collection of $S_i$, such that they cover all steps in $R$ with minimum number of set changes, \ie video source changes.} We use the greedy solution of this problem~\cite{set-covering} to select top-$K$ such set covers. We also use this method in constructing the \emph{distractor set} for retrieval performance evaluation (Sec.~\ref{sec:results}). \KA{We show the effectiveness of this search space reduction in Sec.~\ref{sec:results}.} \JW{See Supp.~for details.}

\begin{table*}[t]\footnotesize
    \centering
    \begin{tabular}{
  L{2.2cm}C{0.4cm}C{0.6cm}C{0.4cm}C{0.6cm}C{0.4cm}C{0.6cm}C{0.4cm}C{0.6cm}|C{0.4cm}C{0.6cm}C{0.4cm}C{0.6cm}|C{0.4cm}C{0.6cm}C{0.4cm}C{0.6cm}   
}
\toprule
\multirow{3}{*}{\raisebox{-0.6em}{Method}}
  & \multicolumn{8}{c|}{Cooking}
  & \multicolumn{4}{c|}{Woodworking}
  & \multicolumn{4}{c}{Gardening} \\
\cmidrule(lr){2-9} \cmidrule(lr){10-13} \cmidrule(lr){14-17}
  & \multicolumn{2}{c}{SaD-MC}
  & \multicolumn{2}{c}{SaD-VD}
  & \multicolumn{2}{c}{HT-Step}
  & \multicolumn{2}{c|}{COIN, CT}
  & \multicolumn{2}{c}{SaD-MC}
  & \multicolumn{2}{c|}{COIN,CT}
  & \multicolumn{2}{c}{SaD-MC}
  & \multicolumn{2}{c}{COIN,CT} \\
\cmidrule(lr){2-3} \cmidrule(lr){4-5} \cmidrule(lr){6-7} \cmidrule(lr){8-9}
\cmidrule(lr){10-11} \cmidrule(lr){12-13}
\cmidrule(lr){14-15} \cmidrule(lr){16-17}
  & MR↓ & R@50 & MR↓ & R@50 & MR↓ & R@50 & MR↓ & R@50
  & MR↓ & R@50 & MR↓ & R@50
  & MR↓ & R@50 & MR↓ & R@50 \\
        \midrule
        CoVR \cite{covr} &193&0.04&132&0.04&161&0.12&97&0.25&29&0.34&37&0.34&31&0.24&26&0.30\\
        VidDetours \cite{detours} &124&0.21&80&0.31&125&0.22&37&0.61&30&0.28&31&0.24&34&0.24&40&0.24\\
        \midrule
        Text-only &108&0.32&76&0.36&123&0.26&78&0.36&44&0.22&31&0.20&51&0.16&58&0.32\\
        InternVideo \cite{internvideo} &36&0.55&8&0.71&68&0.42&19&0.67&38&0.34&43&0.28&45&0.12&45&0.28\\
        Recipe2Video \cite{recipe2video} &125&0.21&50&0.50&93&0.29&26&0.68&40&0.30&75&0.04&48&0.02&76&0.12\\
        \midrule
        Ours &\textbf{3}&\textbf{0.84}&\textbf{3.5}&\textbf{0.91}&\textbf{40}&\textbf{0.56}&\textbf{6}&\textbf{0.88}&\textbf{24}&\textbf{0.36}&\textbf{30}&\textbf{0.38}&\textbf{26}&\textbf{0.36}&\textbf{25}&\textbf{0.36}\\
        \bottomrule
    \end{tabular}
    \vspace{-0.15cm}
    \caption{\textbf{Results on video demonstration retrieval.}  Comparison of the performance of our method against strong \KG{retrieval-based} baselines and prior work using median rank (MR) and recall (R@$50$). The first two methods (CoVR~\cite{covr} and VidDetours~\cite{detours}) are \KG{relevant} retrieval \KG{models}, \KG{though} not specifically designed for this task. Our method significantly outperforms all methods on all metrics, for all test datasets  \KG{and three diverse procedural domains.} \JW{SaD}=\JW{Stitch-a-Demo}. \KG{SaD-VD~\cite{detours} and HT-Step~\cite{htstep-neurips2023} are available only for cooking.} See text.  
    }
    \Major{\KGnote{**are we overlooking a way to quantify the performance of generation methods?  like something crazy where we use a visual similarity metric between the generated frames and the video clip pool and select that way? wondering how we could make ShowHowTo a row in this table.}\KAnote{experiment in progress.... inference is costly with generation...}}
    \label{tab:MergedTable}
    \vspace{-0.25cm}
\end{table*}

\subsection{\modelname~dataset}
\label{sec:dataset}

To train $\mathcal{F}$, we curate a large-scale automatic training data and evaluate the model with a suite of testing data. Each training instance consists of a $(R,V_R)$ pair: a list of \KAnew{procedural} steps and their associated video clips. Positive pairs stem from both real existing \JW{instructional}  
videos as well as novel \JW{procedures} we generate, as described next.  Negatives are alterations of those positives formed as in  Sec.~\ref{sec:model}. 


\textbf{Weakly-supervised training set.} To augment the real positives, the high-level idea is to use all the narrations in the instructional video datasets, along with an existing language model, to create \emph{realistic} \KAnew{weakly-supervised} \KAnew{procedures}
by mixing steps from different demonstrations. This \KAnew{training} data augmentation is essential to help the model learn to combine video clips from different demonstrations.


We have a collection $\mathcal{C}$ containing video demonstrations including HowTo100M~\cite{howto100m}, COIN~\cite{coin}, CrossTask~\cite{crosstask}, or HT-Step~\cite{htstep-neurips2023}. These datasets generally do not have annotations for \KAnew{procedural} steps $R$; only \cite{htstep-neurips2023} has a small scale \KAnew{cooking} recipe description that we use for testing, see below. However, they have narrations that are converted to text using ASR, which we use to obtain \KAnew{procedural} step descriptions. The ASR text is punctuated \cite{tan} and converted to sentences. Then, following~\cite{detours}, we use a language model (Llama-3.1 70B Instruct~\cite{llama-herd}) to convert the narration sentences to step-level, timestamped summaries. Next, we find similar video demonstrations using a sentence similarity score on the summaries (MPNet \cite{mpnet}), and choose pairs, triplets and quartets of \JW{procedures} that have a pair-wise similarity above a threshold $c=0.8$. 

As the last step, we provide 
those summary tuples to the language model and ask it to create a \emph{valid} sequence of steps, 
effectively creating \emph{novel} \JW{procedures}. The question follows the format \emph{``Create a new \JW{procedure} by combining steps from the provided \JW{procedure} summaries. Choose new steps from a different \JW{procedure} only if the current \JW{procedure} cannot be used alone. \JW{Procedures}: ...''}.  See Supp. for the full prompt, which promotes correctness, visual consistency, and accounts for object states.  
The summaries have the time duration of each step, thus the \emph{novel} \KAnew{procedures} use the corresponding start and end time for each video. We show the efficacy of using LLMs in Supp.


Overall, we obtain a weakly-supervised dataset $\mathcal{D}_w = \{(R, V)~|~ R = (r_1, ..., r_n), V = (v_1, ..., v_n), \exists x~\text{s.t.}~v_i = V_x[t_1, t_2], V_x \in \mathcal{C}\}$. See an example in Fig.~\ref{fig:dataset_and_negatives} (right). 
As discussed in Sec. \ref{subsec:task}, 
clips within the same \KAnew{procedure} can come from different source videos. \KAnew{We acknowledge that \KG{\emph{training}} samples in $\mathcal{D}_w$ might contain noise due to the inaccuracies of the LLM. However, the value of unlocking the larger dataset outweighs the risk of introducing noisy training signals, as we show below.  
Manually examining a random sample of labels, we find 75\% to be high quality.}


To reiterate, we use the idea of mixing \KAnew{procedures} (\KG{recipes}) to (a) create a larger training set, and (b) encourage the model to predict 
steps from distinct videos, if needed. \KAnew{The hard negatives, discussed in Sec.~\ref{sec:model}, are sampled from $\mathcal{D}_w$ for contrastive training.} We validate this dataset $\mathcal{D}_w$ with ablations in Supp. 



\textbf{\modelname~testing sets.} 
We create a suite of testing sets to systematically evaluate all aspects of the model design and assumptions.  
Our testing suite has three types of samples: (1) augmented \JW{procedures}  formed using the process above ($w$ for \underline{w}eakly supervised), (2) a manually annotated test set $\mathcal{D}_d$ (for \underline{d}etour) derived from prior work~\cite{detours}, and (3) standard unmixed demonstrations from original videos ($\mathcal{D}_v$, for full \underline{v}ideo).  These three are complementary, offering tradeoffs in realism and strength of ground truth vs.~scale and our control in generating samples.
We describe each in detail next.

Firstly, we have a held-out set from $\mathcal{D}_w$ used as a test set; we call it \textbf{\modelname-MC} set for Mixed Clips. This weakly supervised 
set is large-scale, but naturally incurs some noise (e.g., some step descriptions may not match their associated visual clip due to errors in the LLM summaries).  
While such noise is fine during training, it is a shortcoming for testing, and hence we complement with two more strongly ground-truthed test sets, defined next.

Secondly, we design a test set $\mathcal{D}_d$ called \textbf{\modelname-VD} that strings together the manually verified annotations from VidDetours \cite{detours} to compose a clean test set with minimal manual verification effort.  See Supp. for details. 
 Like $\mathcal{D}_w$, this test set allows evaluating $\mathcal{F}$'s ability to choose clips from different \KAnew{procedures}, but unlike $\mathcal{D}_w$ it is manually verified \KG{and is available only for cooking videos~\cite{detours}}.   

Finally, we use \textbf{standard instructional video datasets} $\mathbf{\mathcal{D}_v}$ (based on CrossTask~\cite{crosstask}, COIN~\cite{coin}, and HT-Step~\cite{htstep-neurips2023}) for testing. We provide the step descriptions for a given video demonstration, and expect the model to \emph{recover} all the video clips from the same video. In cases where the dataset is not specifically annotated for detailed steps (\eg in \cite{coin,crosstask}), we use a language model (similar to $\mathcal{D}_w$ creation) to summarize the ASR text into \JW{procedure} steps. All the output summary steps in these datasets are manually verified for correctness \JW{by us}.  Note that the ground truth clips in these datasets come from the \emph{same} video in $\mathcal{C}$.

In short, sequences in both \modelname-MC and \modelname-VD contain video clips from multiple distinct videos in $\mathcal{C}$. \modelname-MC is auto-created and large-scale, whereas \modelname-VD is manually created and small scale. Meanwhile $\mathcal{D}_v$ is a large-scale source of real (ground truth) videos, but does not require the models to stitch across videos as needed in the ultimate use case.





\begin{figure*}[t] 
    \centering
    \includegraphics[width=1\linewidth]{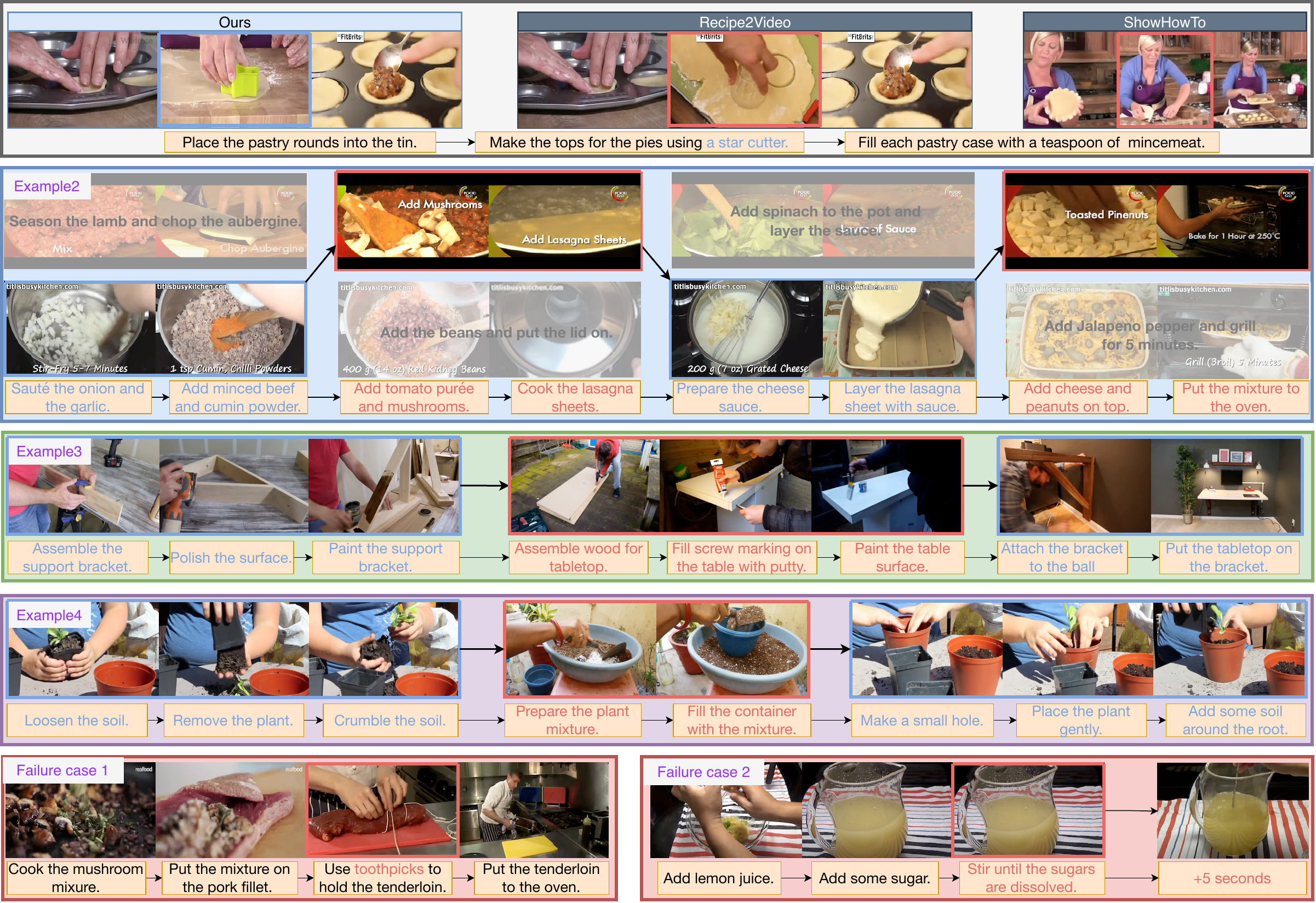} 
    \vspace{-0.25in}
    \caption{\textbf{Qualitative results.} Our method correctly visualizes the step descriptions (top), compared to prior work. 
    \JW{The second to the fourth rows show representative outputs in cooking, woodworking, and gardening.} Our method correctly shows video clips from two video sources. Each of the video sources alone cannot correctly demonstrate all the step descriptions. The last row contains some \textbf{failure cases}, showing the difficulty of the task. Here each keyframe represents a clip $v$; see \JWCam{project page} for actual videos and additional failure analysis. 
    }
    \label{fig:qual} 
    \vspace{-0.25cm}
\end{figure*}

\subsection{Data and implementation details}
\label{sec:implementation}

\textbf{Data sources and dataset statistics.} Sourced from  HowTo100M~\cite{howto100m}, the weakly supervised training set $\mathcal{D}_w$ and the testing set \modelname-MC consist of $446,623$ and $105,742$ samples across cooking, woodworking, and gardening, with $|\mathcal{C}| = 323,177$ and $2,857$, respectively. (See Supp.~for sample counts per domain.) 
The test set SaD-VD derived from VidDetours~\cite{detours} contains $100$ recipes from $235$ unique videos. There are only cooking annotations in~\cite{detours}. Finally, the test data $\mathcal{D}_v$ from COIN~\cite{coin} and CrossTask~\cite{crosstask} $457$ and $942$ across the three domains, and HT-Step~\cite{htstep-neurips2023} contains $1,087$ cooking videos.
HT-Step \cite{htstep-neurips2023} contains step descriptions for cooking videos only from WikiHow \cite{wikihow}, allowing evaluation with original human-written step descriptions. COIN~\cite{coin} and CrossTask~\cite{crosstask} evaluates performance with step descriptions auto-summarized from narrations. 
We 
group COIN and CrossTask into a test set since they are testing the same aspects (see Supp.~for per dataset performance). 


\textbf{Method and training details.}
We use InternVideo2 \cite{internvideo2} as the multi-modal encoder $F_M$. For step localization $F_T$, we use DropDTW \cite{drop-dtw} for its flexibility with extra or missing steps, and its reproducible codebase. The features used in~\cite{mil-nce} are trained on HowTo100M \cite{howto100m}, thus, are used without re-training.
For the \KAnew{procedure} evaluator $F_R$, we use positional encoding and transformer encoder with $4$ layers and $8$ attention heads followed by an MLP layer with $1$ hidden layer. \KAnew{Due to the diverse nature of the domains, } \JW{we train a separate $F_R$ for each of cooking, gardening, and woodworking.} We optimize $F_R$ for $10$ epochs with Adam~\cite{adam} on $8$ Quadro RTX 6000. We set the learning rate as $3\times10^{-4}$ and the batch size as $24$. The $F_M$ output, $F_R$ input, and $F_R$ hidden dimension are all $768$.

\section{Experiments}
\label{sec:results}

We first describe the baselines, testing setup, and metrics and then, the quantitative results. Next, we show the results of a human preference study, followed by ablations and results using our adaptive search space reduction technique. 
 We also show qualitative results, including failure cases.

\noindent\textbf{Baselines.}
We compare our method against strong baselines. We use the same text and video encoder across all baselines, when applicable. All the baselines are trained on the same data as our method for a fair comparison. Each baseline differs in how it selects the sequence of video clips $v_i$ to associate with each step $r_i \in R$.

\begin{itemize}
    \item \textbf{Text-only} computes the average similarity between the ASR transcript of each clip and each step instruction without using the visual cues.
    \item \textbf{InternVideo}~\cite{internvideo2} is a state-of-the-art vision-language model. This baseline computes
    the average similarity between the clip and the step instruction for each step. 
    The baseline uses the same $F_M$ and $F_T$ as our model, but lacks our $F_R$ transformer, \KG{i.e., our procedure evaluator}.
    \item \textbf{Recipe2Video}~\cite{recipe2video} leverages manually-designed metrics that include temporal consistency, information coverage, and cross-modality retrieval to retrieve \JW{procedure} clips.  We reproduced the authors' prior results to ensure correctness of our implementation.
    \item \textbf{CoVR} \cite{covr} is a state-of-the-art method for composed
    video retrieval. 
    Here it retrieves the clip based on the current step and the clip retrieved for the previous step.  
    \item \textbf{VidDetours} \cite{detours} is a state-of-the-art method that we repurpose for our task.  It retrieves the clip based on an original video segment and a user query. For each step, we set the step instruction as the user query and use the previous retrieved clip as the original video segment. We perform this operation sequentially for all \JW{procedure} steps.
\end{itemize}
Further implementation details are available in Supp.

\Major{\KGnote{we need a statement about fairness, e.g., did we use authors' code, how are we showing we gave every possible benefit to the literature baselines (Recipe2Video, CoVR, Detours)?} \KAnote{we mention at the start of the bullets that all of them are trained on the same dataset with same starting features, so I guess that suffice} \KGnote{that's part of it, but also want to refer to how faitfully are we representing their work, e.g. using authors' code? any ways we tried to improve them for our data? because margins are very large so you want to be credible.}\KAnote{will add in the supplementary, about what we used for each of the method and the assumptions.}}

\noindent \textbf{Test setup and metrics.} 
For every test instance, we create a set of $499$ incorrect \emph{distractors} and measure the correct \JW{procedure} retrieval performance out of $500$ samples. 
The distractors contain hard negatives, including samples that violate a constraint,
\KA{and candidates from the adaptive reduced search space (Sec.~\ref{sec:model})}.
Overall, these negative instances are carefully chosen to represent a wide range of possibilities.  See Supp. for details of the negative set construction. 


\begin{table}[t]\footnotesize
    \centering
    \begin{tabular}{L{2.4cm}C{0.9cm}C{0.9cm}C{0.9cm}C{0.9cm}}
        \toprule
         & \multicolumn{4}{c}{Win rate (\%)} \\
        Ours vs & Step & Goal & Quality & Total  \\
        \midrule
        Recipe2Video \cite{recipe2video} &  0.77 & 0.74 & 0.74 & 0.77 \\
        InternVideo \cite{internvideo} &  0.75 & 0.72 & 0.78 & 0.75 \\
        \midrule
        ShowHowTo \cite{showhowto} &  0.94 & 0.94 & 0.85 & 0.98 \\
        \bottomrule
    \end{tabular}
    \vspace{-0.15cm}
    \caption{\textbf{Human preference results.} Our method is preferred by human judges compared to existing retrieval (first two rows) and generation (third row) methods. (Step/Goal: Step/Goal faithfulness, Quality: Visual quality, Total: Overall preference).}
    \label{tab:human_preference}
    \vspace{-0.3cm}
\end{table}

We use standard retrieval metrics \emph{recall@50} 
and the median rank (MR). 
Recall is
higher the better, whereas median rank is lower for better methods. 




\noindent \textbf{Results.} Table \ref{tab:MergedTable} shows the results. Our method significantly outperforms all the strong baselines, on all datasets and metrics \KG{and across all three domains}. Our  gain is up to $29\%$ better in recall and $33$ ranks better in MR, compared to the second-best method, InternVideo~\cite{internvideo}, \KG{which lacks our key innovation, the learned procedure evaluator}. \JWCam{We also compare with HiREST~\cite{Hirest}, a hierarchical retrieval baseline, and our model outperforms it by 0.42 and 0.32 in recall on SaD-VD and HT-Step, respectively.} 


Notably, our method is superior for both when the ground truth video contains demonstrations from multiple source videos, \ie in \modelname-MC and \modelname-VD, as well as in cases where we recover the demonstration from step descriptions, \ie in HT-Step \cite{htstep-neurips2023} and COIN, CrossTask \cite{coin,crosstask}. We attribute this strong performance to our training design that incorporates strong negatives---thus allowing the model to generate output that is \emph{correct} and \emph{visually coherent}.


Fig.~\ref{fig:qual} shows some qualitative results, including failure cases. We see that our method correctly retrieves the clip showing a star shaped cutter, compared to Recipe2Video~\cite{recipe2video} and \KG{the generative} ShowHowTo~\cite{showhowto} (top row). We see our method correctly chooses video clips for given step descriptions, \eg skipping adding beans, and showing tomato purée from another source video (second row). \JW{Our method also successfully retrieves clips for making a wall-mounted painted desk (third row) and preparing a plant mixure for transplanting (fourth row).} Finally, we show some failure cases where the retrieval misses some small objects like a toothpick, showcasing the overall difficulty of satisfying the exact \JW{procedure} steps.

\noindent \textbf{Human preference study.} We also conduct a human preference study to compare \JW{with the strongest retrieval}
\KA{and generation} methods~\cite{showhowto,internvideo,recipe2video}.  
\KAnew{The study considers the cooking subset since typically, we found more people with cooking experience, over gardening and woodworking.}
This complements the results above established with automatic metrics. We use the same settings of InternVideo \cite{internvideo} and Recipe2Video \cite{recipe2video} as above. For ShowHowTo \cite{showhowto},
we follow its original setup and prompt it with the middle frame of the first video clip in the ground truth. All \KG{its} remaining demonstration images are created with the first frame and the step descriptions $R$ as the input.

We evaluate all the methods on four axes---step faithfulness, goal faithfulness, visual quality, and overall preference. 
\Major{\cc{Step faithfulness evaluates which of the two methods shows correct video clips for a given step. Goal faithfulness takes it a step further by assessing if the overall goal of the recipe $R$ is satisfied or not. Visual quality checks for which of the options is easier to watch (with fewer context jumps) and finally, overall preference captures which output the user would prefer.}} 
Every sample is annotated by three subjects unrelated to this project. We compare two methods at a time (one of them always ours), for up to $60$ samples per pair.
See Supp.

Tab.~\ref{tab:human_preference} shows that our method is preferred over all competing approaches, outperforming retrieval-based methods and \KG{the generative} ShowHowTo~\cite{showhowto}. 
This study supports the practical value of our stitching framework: subjects with varying cooking experience (1–10 years) preferred our method 83:17 over \JW{original} video demonstrations, showing that the naturalness of real videos does not offset their limitations in accurately portraying the target multistep task. 

\begin{figure}[t] 
    \vspace{-0.05in}
    \centering
    \includegraphics[width=\linewidth]{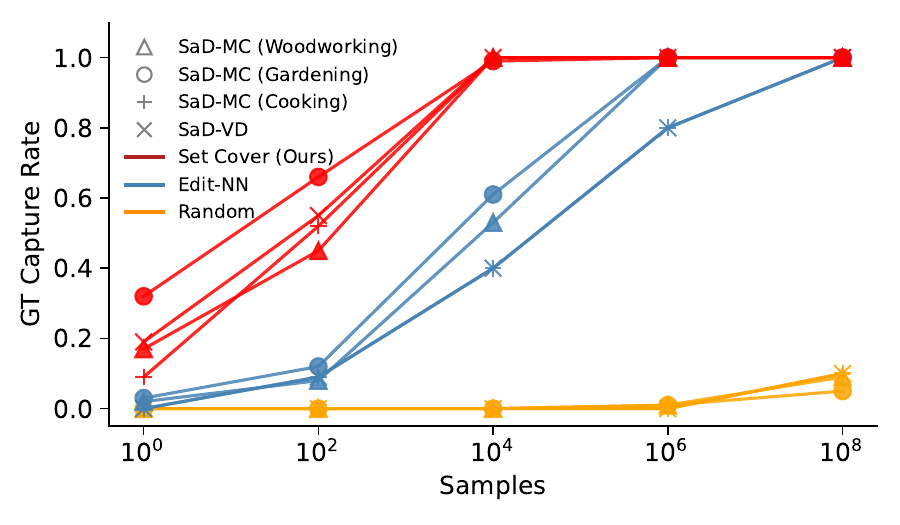} 
    \vspace{-0.37in}
    
    \caption{\textbf{Search space reduction.} Using the effective set cover algorithm, the ground truth (GT) is captured in the candidate set with high probability, even with small sample set sizes. See text. 
    }
    \label{fig:retrieval} 
    \vspace{-0.3cm}
\end{figure}

\noindent \textbf{Adaptive search space reduction.} \JWCam{Fig. \ref{fig:retrieval} shows the percentage of ground truth retrieved across all test cases as a function of the number of retrieved cases per query.} Our set cover algorithm 
\KG{captures} the ground truth in the candidate set with high probability, even for low values of $K$, making it comparable to linear scaling. We see the same trend in \modelname-MC \JW{across three domains} and -VD; 
the algorithm always captures ground truth instances containing only one video source, \ie in 
$\mathcal{D}_v$. We perform better than other methods---randomly selecting clips, and edited-NN, which finds the nearest full video and replaces video clips from other neighbors.
Overall, set cover helps in making our proposed method feasible for real applications.

\vspace{-0.1cm}

\section{Conclusion}
\vspace{-0.1cm}
We propose a novel method \modelname~that 
stitches together video demonstrations that illustrate multistep textual descriptions of procedural tasks. Our method incorporates a novel procedure evaluator, a weakly-supervised large-scale train and clean test data, hard negatives that improve retrieval, and an efficient set cover approach. Our method outperforms strong baselines \JW{up to} $29\%$, 
and human raters prefer our method over SOTA generation and retrieval methods. In the future, we will explore hybrid methods to integrate the retrieved clips with controlled generation, as well as ways to inject preferences into the illustrations such as demonstration speed or language style. 




\section*{Acknowledgement}
\label{sec:acknowledgement}
\JWCam{Research supported in part by the UT Austin IFML NSF AI Institute. We thank all the annotators for their efforts and the lab members in the UT Austin Computer Vision Group for helpful discussions.}
{
    \small
    \bibliographystyle{ieeenat_fullname}
    \bibliography{main}
}

\clearpage
\appendix
\setcounter{page}{1}
\maketitlesupplementary

\section{Supplementary video}

We provide a supplementary video that shows an overview of the paper. It also shows qualitative video examples that best illustrates the visual demonstration we obtain from multistep descriptions.

\section{\modelname~dataset}
\label{sec:supp_dataset}

\textbf{LLM prompts.} As discussed in Sec.~\ref{sec:dataset}, we create weakly supervised training data $\mathcal{D}_w$ for our task. We use a recent language model, Llama 3.1 70B~\cite{llama-herd}, to extract step descriptions from the narrations, and novel \JW{procedure} combinations. We show the LLM prompt for each of the two tasks below:

Firstly, we provide narrations and ask the language model to provide the step descriptions. Here are the prompts for cooking, woodworking, and gardening:

\begin{tcolorbox}[breakable, boxrule=0.2mm]
\textbf{System:} Help summarize the steps of this recipe whose narrations with timestamps are given. Timestamp is given in seconds.

 \textbf{User:} Given the narrations and the timestamp of a video in the format `[start\_time-end\_time] narration text', tell the recipe being made in this video and list down the steps required to complete this recipe. 
 For each step, list down the timestamp of the corresponding narrations that best describe the step. Do not list the introduction, explanation, or comment as a step. Answer in this format: `Recipe: Name of the recipe and brief detail.
 
 Step 1: [start\_time-end\_time] description of the step
 
 Step 2: [start\_time-end\_time] description of the step and so on.'. Here are narrations:

 \textsc{Narration comes here}

 \textbf{Assistant:} 

\end{tcolorbox}

\begin{tcolorbox}[breakable, boxrule=0.2mm]
\textbf{System:} Help summarize the steps of this woodworking project whose narrations with timestamps are given. Timestamp is given in seconds.

 \textbf{User:} Given the narrations and the timestamp of a video in the format `[start\_time-end\_time] narration text', tell the woodworking project being made in this video and list down the steps required to complete this project. 
 For each step, list down the timestamp of the corresponding narrations that best describe the step. Do not list the introduction, explanation, or comment as a step. Answer in this format: `Project: Name of the project and brief detail.
 
 Step 1: [start\_time-end\_time] description of the step
 
 Step 2: [start\_time-end\_time] description of the step and so on.'. Here are narrations:

 \textsc{Narration comes here}

 \textbf{Assistant:} 

\end{tcolorbox}

\begin{tcolorbox}[breakable, boxrule=0.2mm]
\textbf{System:} Help summarize the steps of this gardening project whose narrations with timestamps are given. Timestamp is given in seconds.

 \textbf{User:} Given the narrations and the timestamp of a video in the format `[start\_time-end\_time] narration text', tell the gardening project being made in this video and list down the steps required to complete this project. 
 For each step, list down the timestamp of the corresponding narrations that best describe the step. Do not list the introduction, explanation, or comment as a step. Answer in this format: `Project: Name of the project and brief detail.
 
 Step 1: [start\_time-end\_time] description of the step
 
 Step 2: [start\_time-end\_time] description of the step and so on.'. Here are narrations:

 \textsc{Narration comes here}

 \textbf{Assistant:} 

\end{tcolorbox}

After obtaining the summaries, we feed \JW{$N=2/3/4$} procedures from similar tasks to the language model and ask it to create a novel procedure, following all the desired constraints of realism and correctness. We show the prompt for $N=3$ and for three different domains respectively here:

\begin{tcolorbox}[breakable, boxrule=0.2mm]
\textbf{System:} Generate a new recipe by combining steps from different recipes.

 \textbf{User:} You are tasked with creating a new recipe by combining steps from 3 provided recipe summaries. Your goal is to seamlessly integrate steps from each recipe, switching between them only when necessary due to differences in ingredients, techniques, or tools. Ensure that you do not introduce any new ingredients or steps beyond what is outlined in the summaries. Also, make sure to use at least one step from all recipes. Answer `Not Possible' if this is not possible. Format your response as follows: 
 
 Step 1 (Step \_ in Recipe \_): [Description of the step] 
 
 Step 2 (Step \_ in Recipe \_): [Description of the step]
 
 ... 
 
 Step t (Step \_ in Recipe \_): [Description of the step]
 
 Explanation: [Any explanation that you want to provide] 
 
 Here are the procedures:

 \textsc{Recipe 1 comes here}

  \textsc{Recipe 2 comes here}

   \textsc{Recipe 3 comes here}

 \textbf{Assistant:} 

\end{tcolorbox}

\begin{tcolorbox}[breakable, boxrule=0.2mm]
\textbf{System:} Generate a new woodworking plan by selecting steps from different project instructions.

 \textbf{User:} You are tasked with creating a new woodworking plan by selecting steps from 3 provided project summaries. Your goal is to seamlessly integrate steps from each plan, switching between them only when necessary due to differences in materials, joinery methods, or tools. Ensure that you do not introduce any new materials, techniques, or steps beyond what is outlined in the summaries. Also, make sure to use at least one step from all project summaries. Answer 'Not Possible' if this cannot be done. Format your response as follows: 
 
 Step 1 (Step \_ in Project \_ ): [Description of the step] 
 
 Step 2 (Step \_ in Project \_): [Description of the step] 
 
 ... 
 
 Step t (Step t in Project \_ ): [Description of the step] 
 
 Explanation: [Any explanation that you want to provide] 
 
 Here are the projects:

 \textsc{Project 1 comes here}

  \textsc{Project 2 comes here}

   \textsc{Project 3 comes here}

 \textbf{Assistant:} 

\end{tcolorbox}

\begin{tcolorbox}[breakable, boxrule=0.2mm]
\textbf{System:} Generate a new gardening plan by selecting steps from different project instructions.

 \textbf{User:} You are tasked with creating a new gardening plan by selecting steps from 3 provided project summaries. Your goal is to seamlessly integrate steps from each plan, switching between them only when necessary due to differences in plants, soil preparation methods, or tools. Ensure that you do not introduce any new plants, techniques, or steps beyond what is outlined in the summaries. Also, make sure to use at least one step from all project summaries. Answer 'Not Possible' if this cannot be done. Format your response as follows: 
 
 Step 1 (Step \_ in Project \_ ): [Description of the step]
 
 Step 2 (Step \_ in Project \_): [Description of the step] 
 
 ... 
 
 Step t (Step t in Project \_ ): [Description of the step] 
 
 Explanation: [Any explanation that you want to provide] 
 
 Here are the projects:

 \textsc{Project 1 comes here}

  \textsc{Project 2 comes here}

   \textsc{Project 3 comes here}

 \textbf{Assistant:} 

\end{tcolorbox}

Overall, our language model prompts ensure good-quality weakly supervised data; the effectiveness is also shown in Sec.~\ref{sec:results}.

\textbf{\modelname-VD dataset curation.} In VidDetours's~\cite{detours} manual annotation, for two videos $V_1, V_2 \in \mathcal{C}$, the annotation specifies a time $t$ in $V_1$ where a user asks $Q$ to take a detour to time window $[t_2, t_3]$ in $V_2$. $Q$ is a natural language question like \textit{``Can we substitute garlic with shallots?"}. 
We use $Q$, $t$, and $[t_2, t_3]$ to choose procedure steps from $V_1$ and $V_2$. 
\JW{Since there exists a procedure step $r'$ after time $t_2$ in $V_2$ that answers $Q$, ensuring $r' \in R$ implies $v_i \in V_1$ and $v_{i+1} \in V_2$ for some $i$.} This process can be extended to create a procedure with ground truth containing more videos. Note that the dataset has many distinct annotations with both $V_1 \rightarrow V_2$ and $V_2 \rightarrow V_1$ detours, thus reducing the number of distinct videos in a sample in $\mathcal{D}_d$, see Sec.~\ref{sec:implementation} for statistics. \JWCam{To ensure the correctness of the procedures in $\mathcal{D}_d$, we manually verified text-visual step alignment, object-state consistency, and that the video shows the target procedure.} This process leverages the annotation effort in \cite{detours} to create a clean testing set, with minimal manual verification efforts.

\KA{Note that we cannot use the evaluation dataset in Recipe2Video~\cite{recipe2video} (RecipeQA~\cite{recipeqa} and Tasty videos~\cite{tastyvideos}) since they do not have ground truth annotations; they compare heuristics like abrupt info gain and visual relevance.}

\textbf{Dataset statistics.}
\JW{We provide more detailed statistics for our datasets across three domains. Tab.~\ref{tab:data_count} shows the number of videos in each dataset. Using these videos, Stitch-a-Demo-MC contains 105542, 100, and 100 samples for cooking, woodworking, and gardening. For $\mathcal{D}_w$, it consists of 444823, 900, and 900 samples for the three domains, respectively.} \JWCam{Across different test sets, an average of 53.1 videos with similar content are considered for each sequence, making the test challenging.}
    
\JWCam{We also compare Stitch-a-Demo-MC and Stitch-a-Demo-VD to standard video for object-state consistency using VBench~\cite{vbench} subject consistency at keystep transitions; the curated test sets score only 1.5\% and 1\% lower than standard video, respectively, showing comparable quality.}

\begin{table}[h]\footnotesize
    \centering
    \begin{tabular}{lccc}
        \toprule
        Dataset & Cooking & Woodworking & Gardening  \\
        \midrule
        $\mathcal{D}_w$ & 321139 & 1012 & 1026 \\
        SaD-MC & 2657  & 100  & 100 \\
        SaD-VD & 235 & 0 & 0 \\
        HT-Step~\cite{htstep-neurips2023} & 1087 & 0 & 0 \\
        COIN~\cite{coin} & 337 & 50 & 70\\
        CrossTask~\cite{crosstask} & 812 & 130 & 0  \\
        \bottomrule
    \end{tabular}
    \caption{Number of videos per dataset for cooking, gardening, and woodworking.}
    \label{tab:data_count}
\end{table}



\begin{table*}[t]\footnotesize
    \centering
    \begin{tabular}{
        L{2.3cm}
        C{0.4cm}C{0.7cm}
        C{0.4cm}C{0.7cm}
        C{0.4cm}C{0.7cm}
        C{0.4cm}C{0.7cm}
        C{0.4cm}C{0.7cm}
    }
        \toprule
        \multirow{3}{*}{Method}
        & \multicolumn{4}{c}{Cooking}
        & \multicolumn{4}{c}{Woodworking}
        & \multicolumn{2}{c}{Gardening} \\
        \cmidrule(lr){2-5} \cmidrule(lr){6-9} \cmidrule(lr){10-11}

        & \multicolumn{2}{c}{COIN}
        & \multicolumn{2}{c}{Crosstask}
        & \multicolumn{2}{c}{COIN}
        & \multicolumn{2}{c}{Crosstask}
        & \multicolumn{2}{c}{COIN} \\

        \cmidrule(lr){2-3} \cmidrule(lr){4-5}
        \cmidrule(lr){6-7} \cmidrule(lr){8-9}
        \cmidrule(lr){10-11}

        & MR$\downarrow$ & R@50$\uparrow$
        & MR$\downarrow$ & R@50$\uparrow$
        & MR$\downarrow$ & R@50$\uparrow$
        & MR$\downarrow$ & R@50$\uparrow$
        & MR$\downarrow$ & R@50$\uparrow$ \\
        \midrule

        CoVR \cite{covr}
        & 95 & 0.25
        & 97 & 0.25
        & 33 & 0.82
        & 40 & 0.71
        & 26 & 0.30 \\

        VidDetours \cite{detours}
        & 35 & 0.62
        & 38 & 0.60
        & 46 & 0.54
        & 44 & 0.61
        & 40 & 0.24 \\

        \midrule

        Text only
        & 80 & 0.36
        & 78 & 0.35
        & 31 & 0.76
        & 37 & 0.87
        & 58 & 0.32 \\

        InternVideo \cite{internvideo}
        & 14 & 0.68
        & 23 & 0.66
        & 46 & 0.68
        & 40 & 0.76
        & 45 & 0.28 \\

        Recipe2Video \cite{recipe2video}
        & 25 & 0.70
        & 27 & 0.67
        & 75 & 0.06
        & 76 & 0.07
        & 76 & 0.12 \\

        \midrule

        Ours
        & \textbf{4} & \textbf{0.86}
        & \textbf{4} & \textbf{0.91}
        & \textbf{28} & \textbf{0.92}
        & \textbf{35} & \textbf{0.87}
        & \textbf{25} & \textbf{0.36} \\

        \bottomrule
    \end{tabular}
    \caption{\textbf{Additional results on visual demonstration retrieval.} Comparison of the performance of our method against baselines and prior work for COIN~\cite{coin} and CrossTask~\cite{crosstask}. This table shows per dataset performance for the consolidated results in Tab.~\ref{tab:MergedTable}.}
    \label{tab:coin_crosstask}
\end{table*}

\begin{figure}[t] 
    \centering
    \includegraphics[width=0.8\linewidth]{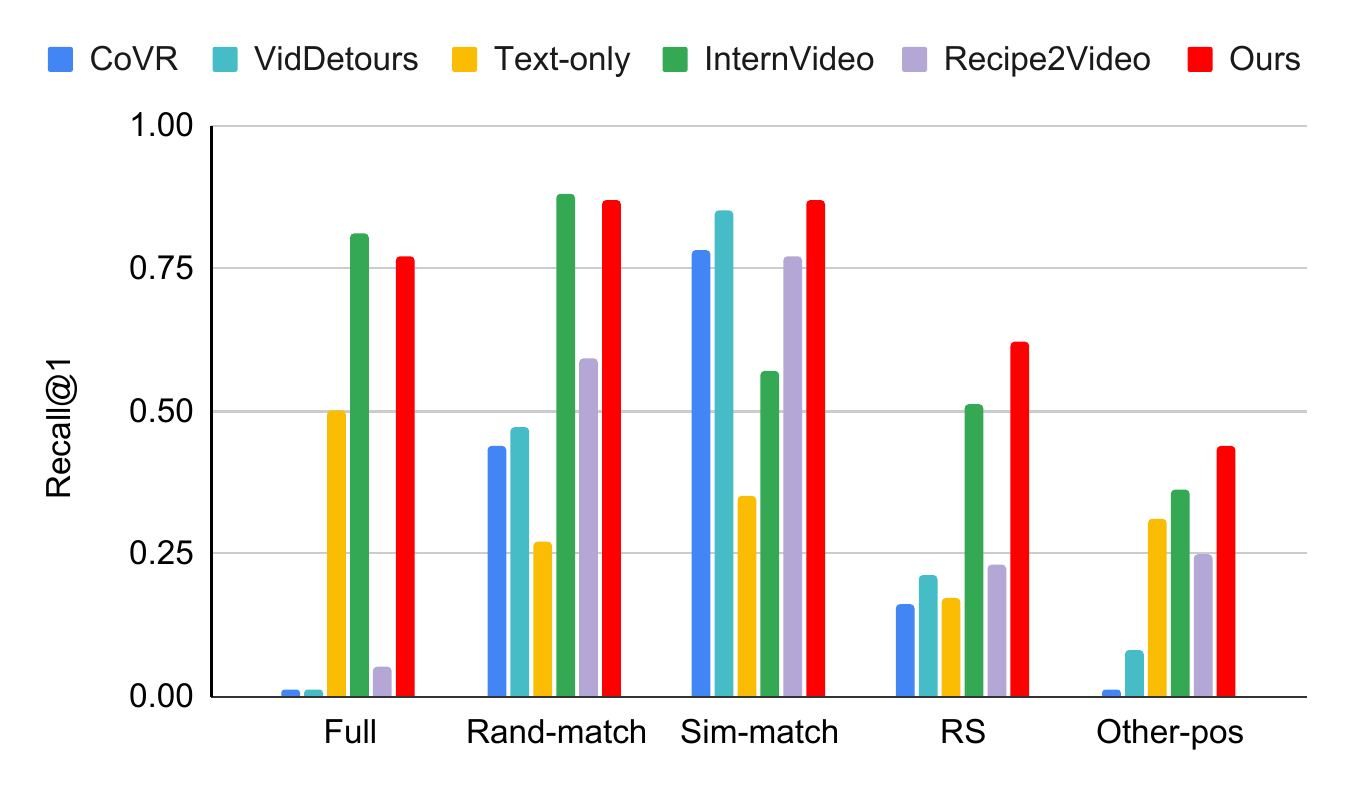} 
    \caption{\textbf{Result on distractor set splits}. Our model performs competitively on all splits---particularly the more challening RS, Other-pos, and Sim-match.
    }
    \label{fig:onehundredsplit} 
    \vspace{-0.3cm}
\end{figure}

\begin{table}[t]\footnotesize
    \centering
            \begin{tabular}{L{3.7cm}C{0.6cm}C{0.6cm}C{0.6cm}}
            \toprule
            Method & MedR↓ & R@1↑ & R@5↑ \\
            \midrule
            w/o augmentation  & 70  & 0.03  & 0.14  \\
            Temporally-sampled procedures & 10 & 0.18 & 0.42 \\
            Weakly supervised $\mathcal{D}_w$ (ours)    & \textbf{3.5}   & \textbf{0.23}  & \textbf{0.56}  \\
            \bottomrule
        \end{tabular}
\vspace{0.2cm}

    \begin{tabular}{l|c|c|c|c|c}
        \hline
        \textbf{Cor} & \textbf{Con} & \textbf{OSC} & \textbf{MedR↓} & \textbf{R@1↑} & \textbf{R@5↑} \\
        \hline
        \checkmark &   &   & 9   & 0.17 & 0.41\\
         & \checkmark   &   & 171  & 0 & 0.03  \\
        &  &  \checkmark & 11   & 0.11 & 0.39 \\
        \checkmark & \checkmark &   & 9   & 0.18 & 0.45 \\
        & \checkmark & \checkmark & 15 & 0.07 & 0.21 \\
        \checkmark &   & \checkmark &  \textbf{5}   & 0.15 & \textbf{0.54} \\
        \checkmark & \checkmark & \checkmark & \textbf{5}  & \textbf{0.22} & 0.52 \\
        \hline
    \end{tabular}
    \vspace{-0.15cm}
    \caption{\textbf{Ablations.} Our weakly-supervised data is effective for training, compared to alternatives (top). All the constraints for hard negatives are useful (bottom). (Cor: step/goal correctness, Con: visual continuity, OSC: object state consistency.) 
    }
    \label{tab:ablations}
    \vspace{-0.5cm}
\end{table}

\section{Results and additional ablations}
\label{sec:results_in_supp}

\textbf{Baseline implementation details.} We use the implementation provided in \cite{tan} to extract InternVideo~\cite{internvideo} visual features. We use the author's provided code for VidDetours~\cite{detours} and CoVR~\cite{covr}, and retrain on our dataset for a fair comparison. We re-implement Recipe2Video~\cite{recipe2video} since there is no publicly available codebase. \JW{We matched our re-implementation on RecipeQA \cite{recipeqa} and obtain a visual relevance of 0.82, better than the reported 0.8, establishing the correctness of our usage.}

\noindent \textbf{Creating the distractor set for evaluation.} In Sec.~\ref{sec:results}, we state that we create a \emph{distractor set} containing $499$ negative samples for retrieval performance evaluation. We describe the composition of this challenging negative set, that assesses various aspects of the visual demonstration creation. The dataset consists of equal samples from the following strategies:
\begin{itemize}
    \item Reduced search space top-K samples (RS): We use the set cover algorithm, as introduced in Sec.~\ref{sec:model}, to create a set of challenging options. These options contain visual demonstrations that combine multiple source videos.
    \item Full videos (Full): We sample videos from the same task that serves as an unmixed distractor in the candidate set.
    \item Other positives (Other-pos): These are ground truth visual demonstrations for other procedures in the test set.
    \item Random mix-n-match (Rand-match): These distractors are a random combination of video clips from the same task.
    \item Mix-n-match w/ similarity (Sim-match): For every step description, we choose the highest similarity visual demonstrations and create a sequence from top retrievals, based on the similarity scores. Note that these options do not consider the visual continuity, unlike our training positives.
\end{itemize}

Overall, \JWCam{we carefully design negatives in the distractor set} that serve as a competitive benchmark for this task.

\begin{figure*}[t]
    \centering
    \includegraphics[width=\linewidth]{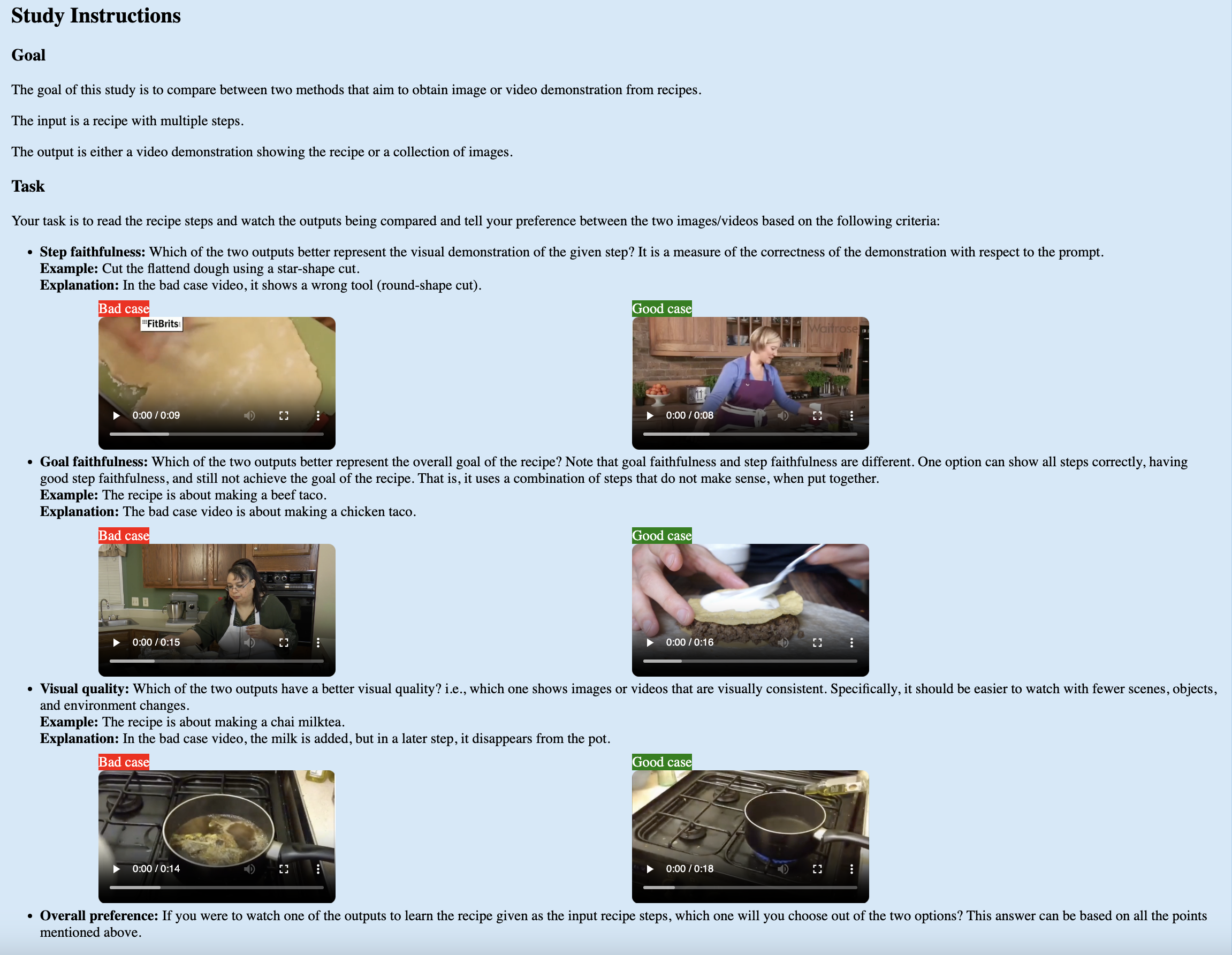}
    \caption{\textbf{Human preference study interface instructions.} We provide examples of all axes for human preference study---step faithfulness, goal faithfulness, visual quality and the overall preference.}
    \label{fig:human_eval_instructions}
    \vspace{-0.15in}
\end{figure*}

\textbf{Additional result splits.} We provide performance splits for results presented in Sec.~\ref{sec:results}. We show results for the following splits:

\begin{itemize}
    \item Performance split between COIN and CrossTask: Tab.~\ref{tab:coin_crosstask} shows performance split between COIN~\cite{coin} and CrossTask~\cite{crosstask}. We show the combined performance in Sec.~\ref{sec:results} since both the datasets evaluate the same aspects of the model. We outperform all the baselines on all for metrics for both the datasets. \JW{Note that CrossTask \cite{crosstask} does not have gardening videos.}
    \item Performance across various distractor set components: In Sec.~\ref{sec:results_in_supp}, we introduce the distractor set components. We evaluate the performance with each component of the distractor set. For example, we evaluate the retrieval performance with $99$ negative samples from `Random mix-n-match' and one ground truth. Fig.~\ref{fig:onehundredsplit} shows the results. Our method consistently performs good on all distractor sets. In particular, our method outperforms all the competing methods on more difficult distractor sets, \ie, \KA{RS, Sim-match, and Other-pos. }
\end{itemize}

\noindent \textbf{Ablations.} 
We present two key ablations:


\emph{Weakly-supervised dataset creation}: In Sec.~\ref{sec:dataset}, we introduce a pipeline to create weakly-supervised training data $\mathcal{D}_w$. This method used an LLM to create novel procedures, and the corresponding video demonstrations. We compare the effectiveness of this data augmentation, compared to 
(a) w/o augmentation, where we train only with videos in $\mathcal{C}$ (\ie HowTo100M \cite{howto100m}) directly, and (b) temporally-sampled procedure combination, where we use videos from the same task and create new procedures by combining steps happening at the same time. The second method also creates novel procedures, however, unlike our approach, it does not leverage an LLM to create \emph{plausible} transitions. Tab. \ref{tab:ablations} shows the results on \modelname-VD test set. 
Our method clearly outperforms both the data creation alternatives, showing the effectiveness of our LLM-aided weakly supervised set $\mathcal{D}_w$. \KA{We see the same trend in all the test datasets.}

\emph{Negative sampling strategy}: We introduce negative sampling strategies in Sec.~\ref{sec:model}. We compare the effect of individual negative samples in Tab.~\ref{tab:ablations} (bottom). We observe that all three of \emph{step correctness}, \emph{visual continuity}, and \emph{object state continuity} are crucial for our method's performance.

\JWCam{
\noindent \textbf{More fine-grained analysis.}
As in \ref{sec:supp_dataset}, we leverage VBench \cite{vbench} subject consistency at keypoint transitions as a fine-grained measure of procedure retrieval quality. Our method scores 0.67, outperforming the strongest baseline, InternVideo \cite{internvideo2}, which scores 0.62, demonstrating better object-state consistency in the retrieved clips.
}

\begin{figure*}[t]
    \centering
    \includegraphics[width=\linewidth]{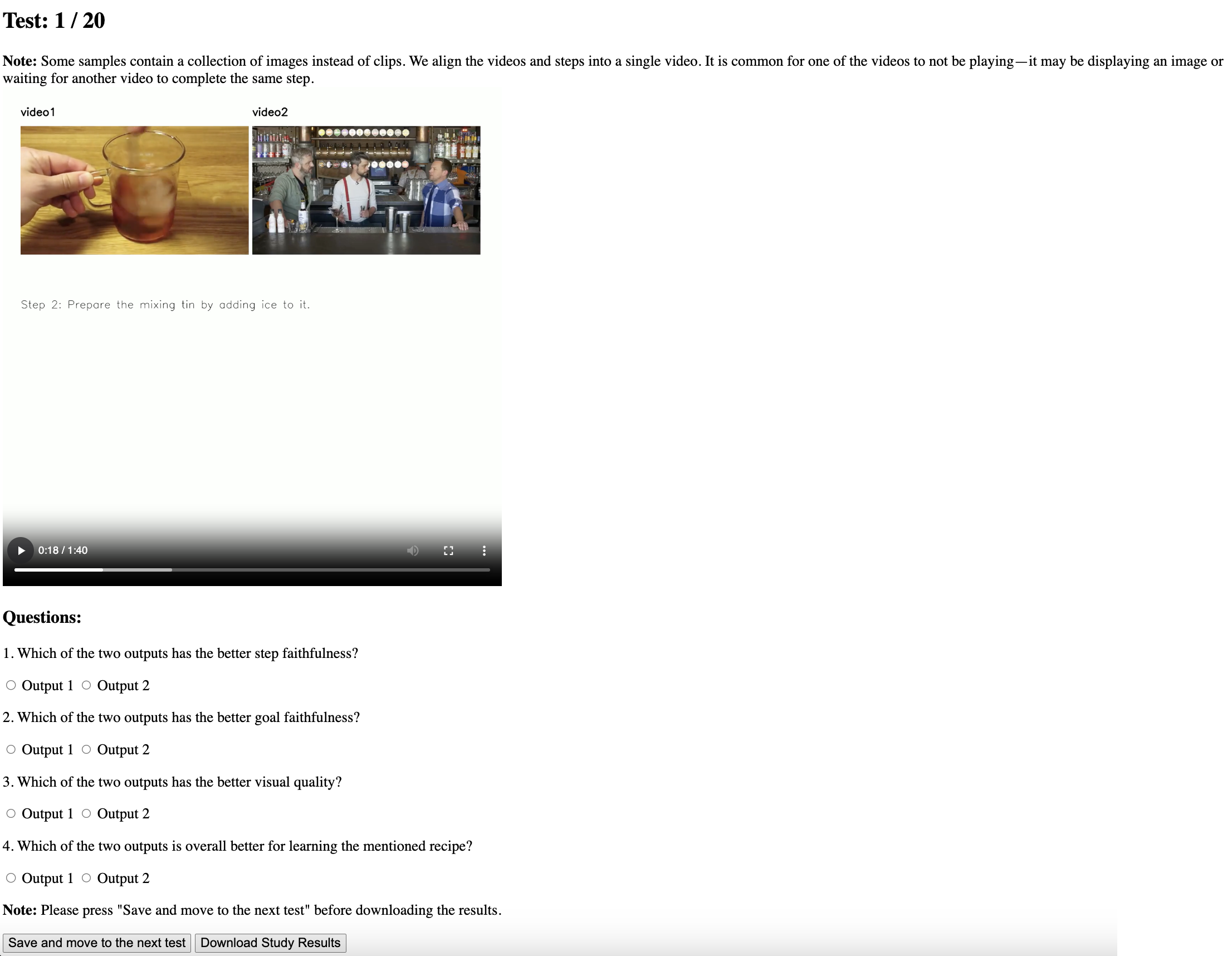}
    \caption{\textbf{Human preference study submission form.} The video in the interface shows both the candidate procedures side by side, and the step description is shown below. The video is followed by four questions, asking about each axis, and the result is saved as a CSV file.}
    \label{fig:human_eval_questions}
    \vspace{-0.3cm}
\end{figure*}

\section{Adaptive search space reduction}
In Sec.~\ref{sec:model} we introduce our set cover algorithm that reduces the search space for practical deployment. For computational complexity, given a procedure query with $M$ steps and a pool of $N$ videos with $K$ clips each, our method selects the top $S$ clips per step in $O(NK \text{ log } S)$ time and explores coherent combinations in $O(SM^{2})$, finishing at combinations that minimize switches between sources.

\section{Human preference study interface}

\JWCam{We invited 20 annotators for evaluation, with each comparison annotated by three people. The annotators show high consistency (85.2\% pairwise agreement)}. Fig.~\ref{fig:human_eval_instructions} and Fig.~\ref{fig:human_eval_questions} show our designed human preference study interface instructions and questions, respectively. We first provide instructions, followed by examples of all aspects we evaluate---step faithfulness, goal faithfulness, visual quality and overall preference (Fig.~\ref{fig:human_eval_instructions}).
\KA{Step faithfulness evaluates which of the two methods shows correct video clips for a given step. Goal faithfulness takes it a step further by assessing if the overall goal of the procedure $R$ is satisfied or not. Visual quality checks for which of the options is easier to watch (with fewer context jumps) and finally, overall preference captures which output the user would prefer.}
The two options are randomly shuffled to avoid any bias in the human subjects. All the subjects are unrelated to this project and briefed about the task before the study. 

\end{document}